\def\year{2019}\relax
\documentclass[letterpaper]{article} 
\usepackage{aaai19}  
\usepackage{times}  
\usepackage{helvet}  
\usepackage{courier}  
\usepackage{url}  
\usepackage{graphicx}  
\usepackage{multirow}
\usepackage{amsmath,amssymb} 
\begin{document}
%
\title{Keypoint Based Weakly Supervised Human Parsing}
\author{Zhonghua Wu, Guosheng Lin, Jianfei Cai \\
Nanyang Technological University, Singapore\\
\{zhonghuawu, gslin, asjfcai\}@ntu.edu.sg
}
\maketitle

\begin{abstract}

Fully convolutional networks (FCN) have achieved great success in human parsing in recent years. In conventional human parsing tasks, pixel-level labeling is required for guiding the training, which usually involves enormous human labeling efforts. To ease the labeling efforts, we propose a novel weakly supervised human parsing method which only requires simple object keypoint annotations for learning. We develop an iterative learning method to generate pseudo part segmentation masks from keypoint labels. With these pseudo masks, we train a FCN network to output pixel-level human parsing predictions. Furthermore, we develop a correlation network to perform joint prediction of part and object segmentation masks and improve the segmentation performance. The experiment results show that our weakly supervised method is able to achieve very competitive human parsing results. Despite our method only uses simple keypoint annotations for learning, we are able to achieve comparable performance with fully supervised methods which use the expensive pixel-level annotations.

\end{abstract}

\section{Introduction}

Semantic image segmentation is a fundamental task for image understanding.
Human parsing, also known as human part segmentation, can be considered as a part-level image segmentation task.
Human parsing aims to segment one person into different parts, which is a pixel labeling task and plays an important role in human analysis.
Part segmentation or human parsing has recently attracted increasing attention in the research community \cite{Lin:2017:RefineNet,chen_cvpr14,liang2015deep,wang2015joint,xia2016zoom}.
Human parsing stimulates various high-level vision understanding applications such as action recognition, human behavior analysis and video surveillance.

Conventional part segmentation methods require pixel-level annotations for training， which usually involve excessive human labeling efforts.

\begin{figure*}[t]
  \center
\includegraphics[width=0.8\textwidth]{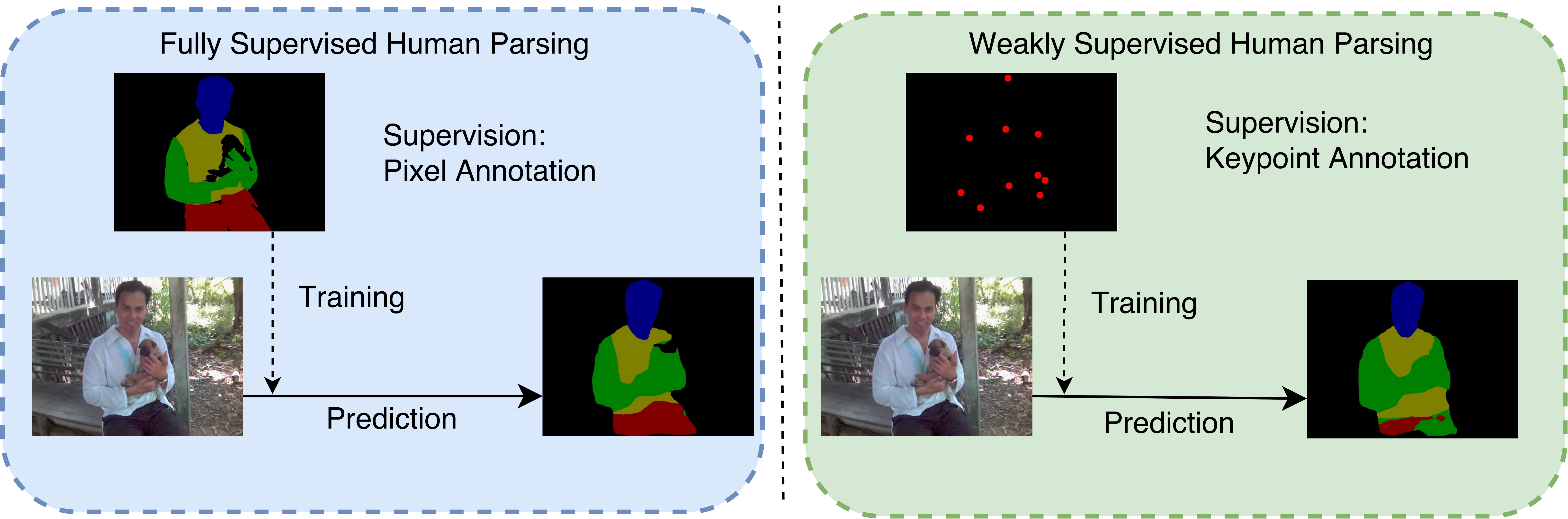}
\caption{Comparison between fully supervised methods and our weakly supervised method. The left box (blue) describes traditional fully supervised human parsing,
which requires expensive pixel-level annotations for training. The right box (green) illustrates our weakly supervised human parsing method which only requires simple object keypoint annotations for training. We are able to achieve comparable performance with fully supervised methods. }
\label{weakly}
\end{figure*}

To avoid this huge burden of pixel-level annotations, in this research we propose to use simple object keypoint annotation as supervision for learning human parsing models.
Compared to pixel-wise part annotations, object keypoint annotations are much easier to obtain which significantly reduces human labeling efforts. Fig. \ref{weakly} illustrates the idea of our keypoint based weakly supervised human parsing framework and how it is different from conventional fully supervised methods.

Object keypoint annotations can be obtained from human labelling, e.g., keypoint annotations in human pose datasets \cite{xia2017joint,lin2014microsoft}, or from pre-trained human pose estimation models, e.g, Mask RCNN \cite{he2017mask} and AlphaPose \cite{fang2017rmpe,2018arXiv180200977X}.
We demonstrate our proposed method is able to incorporate with human labeled keypoint annotations as well as the less accurate keypoint predictions generated by pre-trained human keypoint detection models
to achieve pixel-level human part and object segmentation.
In addition, considering there is a strong correlation between the whole object segmentation and the part segmentation. We develop a joint learning method to model such correlations and simultaneously output object and part segmentation masks.
Particularly, we introduce a correlation block to model interaction between the part prediction and the object prediction, and it helps to improve the final part segmentation result.

Our main contributions are summarized as follows:

\begin{itemize}
  \item We propose a weakly supervised method to ease the human labeling efforts for human parsing. We are able to achieve good pixel-level human part and object segmentation results using only simple object keypoint annotations as supervision for learning. The object keypoints can be obtained from human manual labeling or pre-trained object keypoint detectors. Our method significantly reduces human labeling efforts and achieves very competitive performance for human parsing.
  \item We propose an iterative learning method to generate accurate pseudo masks for parts and objects from object keypoint annotations. With such high-quality pseudo masks, we train a segmentation network to jointly predict pixel-level part and object segmentation masks.
  \item Due to the strong correlation between parts and objects, joint prediction is expected to benefit the segmentation performance. We propose a correlation network to simultaneously output part and object segmentation masks, and achieve improved results for part segmentation.

\end{itemize}

\section{Related work}

Our method is related to the research themes including weakly supervised segmentation,
human part segmentation and pose estimation.

\subsection{Weakly supervised segmentation}

In the recent years, the development of deep convolutional neural networks (CNN) with advanced
network structures such as VGG \cite{DBLP:journals/corr/SimonyanZ14a} and ResNet \cite{he2016deep} have been widely used in many areas such as
object detection and segmentation.
The work in \cite{long2015fully} proposes fully convolutional neural networks (FCN) based on VGG network for semantic segmentation with end-to-end learning.
The approach in \cite{chen2016deeplab} introduces Atrous/dilated convolution and employs fully connected CRFs to improve the FCN method.

Conventional fully supervised segmentation requires pixel-wise mask annotations for training and it requires enormous human labeling effort which is usually excessively expensive.
To ease the labeling efforts, a number of weakly supervised methods  \cite{pathak2014fully,vernaza2017learning,lin2016scribblesup,bearman2016s,dai2015boxsup,PC2015Weak,ShenBMVC17,zhangweakly2018}
have been proposed to employ weak supervision for learning segmentation models.
For example, the methods in \cite{pathak2014fully,PC2015Weak,ShenBMVC17} use image level labels for learning
segmentation models;
the work in \cite{bearman2016s} use image level and point level information for model training.
The work in \cite{dai2015boxsup} and \cite{lin2016scribblesup} use box annotation and scribble annotation, respectively, as supervision for learning segmentation models.

Different from these existing studies, we focus on human part segmentation. We propose to use object keypoint annotation as supervision which is more challenging than using scribble or box-level supervision for learning segmentation models.

\subsection{Human part segmentation }

The work in \cite{chen_cvpr14} extends object segmentation to object part-level segmentation.
It releases a PASCAL PART dataset which contains pixel-level part annotations.
The work in \cite{wang2015joint} first attempts part segmentation on animals.

It uses fully-connected CRFs as post-processing to enhance the consistency between parts and objects.

The approach in \cite{CY2016Attention} proposes an attention model to fuse multi-scale prediction for part segmentation.
The work in \cite{xia2016zoom} uses the ``auto-zoom" to build a hierarchical model to adapt the scales for objects and parts.

The method in \cite{li2017towards} extends single human parsing to multiple human
parsing and demonstrates in real-world applications.
The method in \cite{tsogkas2015deep} employs high-level information to improve
part segmentation.
The approach in \cite{Lin:2017:RefineNet} proposes a multi-path refinement network to achieve high resolution and accurate part segmentation.

Different from these fully supervised works, we focus on weakly supervised human parsing which only use human keypoint annotations rather than pixel level annotations.

\subsection{Pose estimation}

The work in \cite{cao2017realtime} proposes Part Affinity Fields for human pose estimation.
Then multi-stage pose estimation methods \cite{fang2017rmpe,he2017mask}
use object detection and segmentation information to guide pose estimation prediction.
The method in \cite{xia2017joint} jointly performs the multi-person pose estimation and the semantic part segmentation in a single image to enhance the performance of both segmentation and pose estimation.

In this work, we use human pose estimation method to generate pseudo human keypoint to refine or generate our human part segmentation.

\begin{figure*}[t]
  \center
\includegraphics[width=0.8\textwidth]{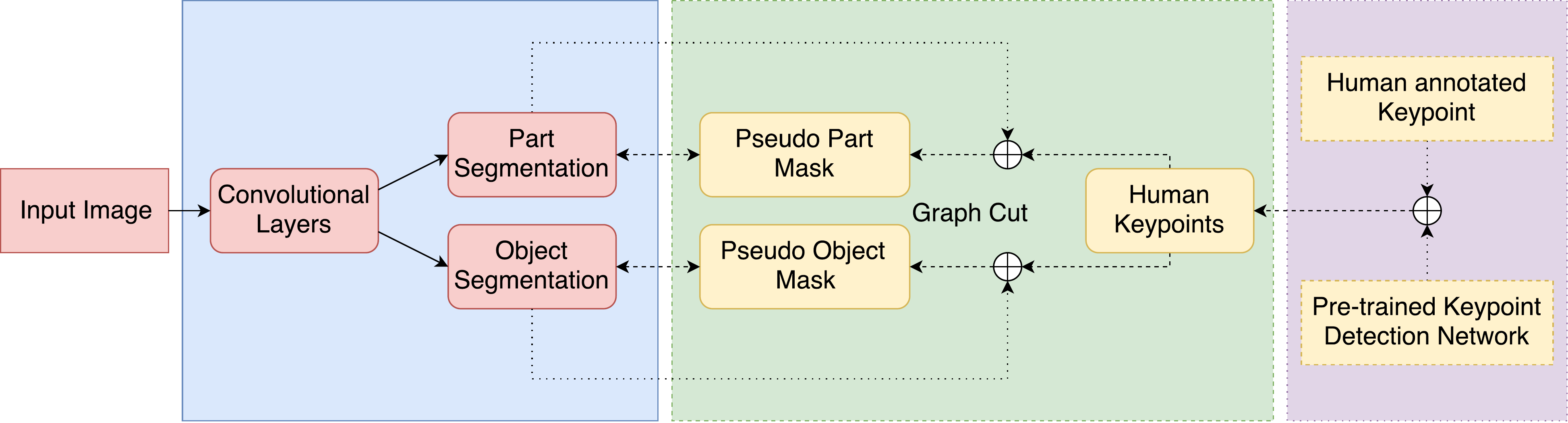}
\caption{An overview of our weakly supervised method.
The right part (purple) shows the object keypoint annotations which can be obtained from human labeling or pre-trained keypoint detectors such as MaskRCNN or AlphaPose.
The middle part (green) illustrates our pseudo mask generation for parts and objects from keypoint annotations.
The left part (blue) describes our FCN based segmentation network with two output branches for a joint object and part learning. This segmentation network uses pseudo masks described in the middle part for training.
In the test stage, only the trained segmentation network (left part) is applied for the part segmentation prediction.
}

\label{overall}
\end{figure*}

\section{Approach}

Fig. \ref{overall} gives an overview of the proposed method. It consists of three parts: the human keypoint annotation part in the right, the pseudo mask generation in the middle, and the FCN based segmentation network in the left. In particular, we generate
pseudo object and part masks from object keypoint annotations.
Then the resulting pseudo masks are employed for training our FCN based joint segmentation networks for part and object mask prediction. In the following, we focus on elaborating our three specially designed processes, i.e. generating pseudo marks from keypoints, iteratively refining pseudo marks and correlation network for joint prediction.

\subsection{Generating pseudo masks from keypoints}
\label{Generate pseudo mask from keypoints}

\begin{figure}[t]
\center
\includegraphics[width=0.45\textwidth]{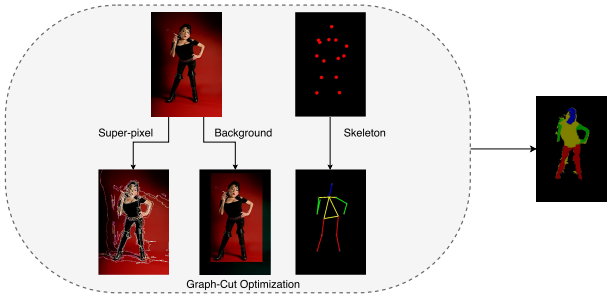}
\caption{
Illustration of the pseudo mask generation from keypoint annotations.
Firstly, we generate super-pixels \cite{uijlings2013selective} of the input image (first bottom column).
Secondly, we estimate background regions based on the location of keypoints (second bottom column). We treat the pixels which are 50 pixels far away the human keypoints as the background.
Thirdly, we connect the object keypoints to generate the skeleton (third bottom column).
Finally, we construct a graph-cut model to generate the pseudo masks of parts.}
\label{graph cut}
\end{figure}

\begin{figure*}[t]
\center
\includegraphics[width=.8\textwidth]{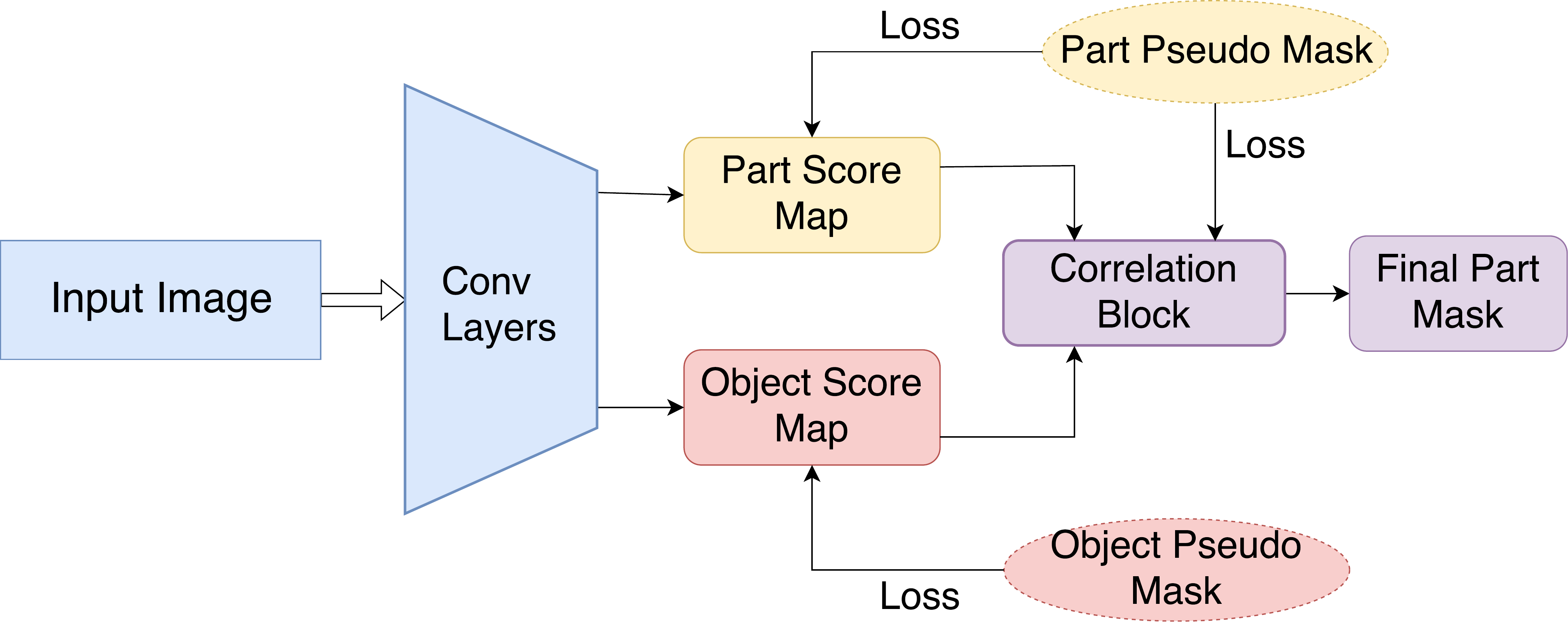}
\caption{Illustration of our correlation network for joint learning of parts and objects.
Our network contains two branches for part and object prediction.
We introduce a correlation block to model the interaction between parts and objects, and thus to improve the final part segmentation.}
\label{structure loss}
\end{figure*}

\begin{figure}[t]
\center

\includegraphics[width=.45\textwidth]{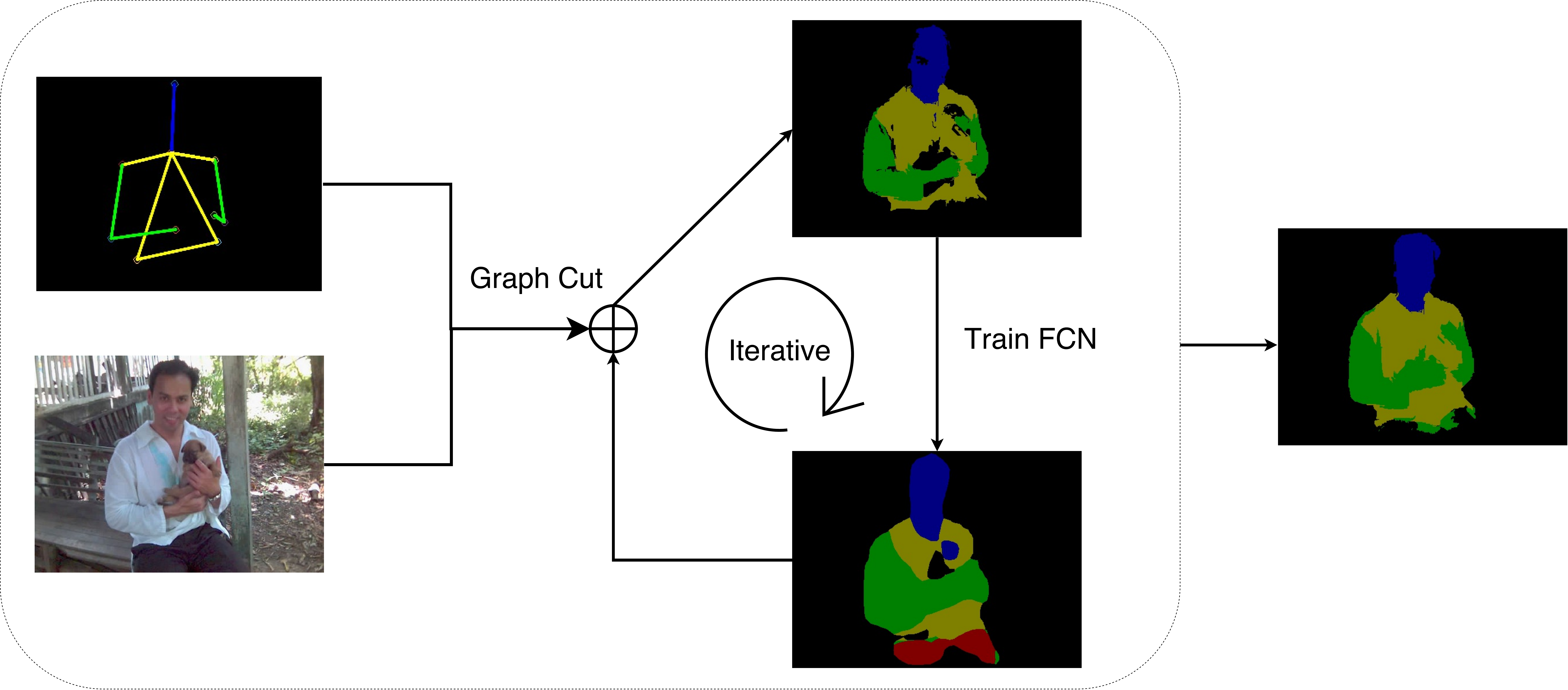}
\caption{Illustration of our iteration refinement process for pseudo mask generation.
We perform graph cut prediction and FCN training iteratively to improve pseudo masks.
In the first iteration, we build a graph-cut model based on object keypoints to generate pseudo masks, and then we train FCN based segmentation network using the pseudo masks.
In the next iteration, we jointly consider the keypoint annotations and the segmentation score map, generated by the trained FCN based model from the last iteration, to construct a new graph-cut model to construct better pseudo masks. We repeat a few iterations to output the final segmentation. }
\label{iter_train}
\end{figure}

At the first step, we build a graphical model to generate pseudo masks of objects and parts from keypoint annotations.
We generate super-pixels for our training images and construct a graph over the super-pixels.
This problem can be formulated as an energy minimization problem. The energy function is written as:

\begin{equation}
E = \sum_i \varphi_i ( y_i ) +  \sum_{i,j} \varphi_{ij} ( y_i, y_j ) ,
\label{energy}
\end{equation}
where $ \varphi_i ( y_i ) $ is the unary term indicating the labeling confidence for one super-pixel,
and $ \varphi_{ij} ( y_i, y_j ) $
is the pairwise term indicating the pairwise labeling confidence for a pair of neighboring super-pixels.
Here $ y \in \{1 ... K\} $ denotes the part label which takes a value from one of the $ K $ labels.

We construct the unary term based on object keypoint annotations.
As shown in Fig. \ref{graph cut}, we connect object keypoints to generate a skeleton, and all pieces of the skeleton are assigned object part labels based on the types of object keypoints.
A super-pixel overlapped with the skeleton will be assigned a part label, denoted by $L$, according to the overlapped skeleton pieces.
If there are two skeletons across one super-pixel, the superpixel will be given the label of the skeleton with more overlapped pixels.
We consider all part regions as confident foreground regions.
We formulate the unary term cost function as:

\begin{equation}
\varphi_i ( y_i ) =
\begin{cases}
-log(\frac{1}{\mid K \mid})  & if X_i \cap S = \varnothing  ; \cr
0                            & if X_i \cap S \ne \varnothing , y_i  =  L_i  ; \cr
a \ large \ value                & if X_i \cap S \ne \varnothing , y_i \ne L_i ,
\end{cases}
\label{graph_cut}
\end{equation}
where $ X_i $ indicates a super-pixel and $S$ indicates the skeleton.
Here $ X_i \cap S = \varnothing $ indicates a super-pixel $ X_i $ does not overlap with any pieces of the skeleton $ S $, and likewise, $ X_i \cap S \ne \varnothing $ indicates a super-pixel $ X_i $ overlaps with some pieces of the skeleton $ S $.
In the first case, i.e. the super-pixel does not overlap with the skeleton, the costs of all part categories are set to the same, with $ K $ indicating the total number of part categories.

Recall we assign a part label, denoted by $ L_i $,  to the super-pixel $i$ based on the overlapped piece of the skeleton.
In the cases when the super-pixel overlaps with the skeleton, if $y_i$ equals to the assigned label $L_i$, we set the cost to 0; otherwise, we set a high cost value, e.g., $10^7$ in our implementation.

We build the pairwise term to model the local smoothness information.
Following the work in \cite{boykov2004experimental}, we construct the pairwise term for a pair of neighboring super-pixels based on color, position and texture information, denoted by subscript $ C $, $ P $ and $ T $, respectively. The pairwise term can be written as:

\begin{equation}
\begin{split}
\varphi_{ij} ( y_i, y_j ) = &\omega_C \exp(-\frac{\parallel h_C(x_i) - h_C(x_j) \parallel^2 }{2 \sigma_C^2}) \\
&+ \omega_P \exp(-\frac{\parallel h_P(x_i) - h_P(x_j) \parallel^2 }{2 \sigma_P^2}) \\
&+ \omega_T \exp(-\frac{\parallel h_T(x_i) - h_T(x_j) \parallel^2 }{2 \sigma_T^2}),
\end{split}
\end{equation}
where $ h_C $ $ h_P $ and $ h_T $ are the histogram features for the color, position and texture respectively, $ \omega_C $ , $ \omega_P $ and $ \omega_T $ are the trade-off parameters of different terms.
Here $ \sigma_C $ , $ \sigma_P $ and $ \sigma_T $ are the bandwidth parameters. We use the multi-label graph cut to minimize the energy function \cite{boykov2004experimental}.

\subsection{Iterative refinement of pseudo masks}
\label{Iterative refinement for pseudo mask}

As shown in Fig. \ref{iter_train}, with the resulting pseudo part segmentation masks, we train an FCN based part segmentation model.
We use DeepLab \cite{CY2016Attention} segmentation method as our based model with VGG \cite{simonyan2014very} as our base network.
The trained FCN model is applied to generate the final part segmentation.

We obtain part score maps from the trained FCN model.
The part score maps can be incorporated into the unary term in Equation \ref{energy} to further improve the pseudo mask generation. In particular, the energy function Equation in \ref{energy} can be updated as:

\begin{align}
E = \sum_i \varphi_i^S ( y_i ) +  \sum_i \varphi_i^N ( y_i ) + \sum_{i,j} \varphi_{ij} ( y_i, y_j ),
\end{align}
where $ \varphi_i^S $ is same as the unary term from Equation \ref{energy}, which is constructed based on the skeleton information,
$ \varphi_i^N ( y_i ) $ is based on the FCN part score map,
and $ \varphi_{ij} ( y_i, y_j ) $ is the same as the pairwise term in Equation \ref{energy} to model the local smoothness of neighboring super-pixels.

\begin{table*}[t]
  \caption{Information of object keypoint annotations in two datasets
  }
\centering
\resizebox{0.8\linewidth}{!}{
\begin{tabular}{|l|l|l|l|l|l|}
\hline
                    & head                                                    & torso                                                         & arm                                                              & leg                                                        & object          \\ \hline
Human keypoint in PASCAL& \begin{tabular}[c]{@{}l@{}}Forehead\\ Neck\end{tabular} & \begin{tabular}[c]{@{}l@{}}Neck\\ Shoulder / Hip\end{tabular} & \begin{tabular}[c]{@{}l@{}}Shoulder\\ Elbow / Wrist\end{tabular} & \begin{tabular}[c]{@{}l@{}}Hip\\ Knee / Ankle\end{tabular} & All Joint Point \\ \hline
Human keypoint in COCO & \begin{tabular}[c]{@{}l@{}}Nose\\ Eye / Ear\end{tabular} & \begin{tabular}[c]{@{}l@{}}Neck\\ Shoulder / Hip\end{tabular} & \begin{tabular}[c]{@{}l@{}}Shoulder\\ Elbow / Wrist\end{tabular} & \begin{tabular}[c]{@{}l@{}}Hip\\ Knee / Ankle\end{tabular} & All Joint Point \\ \hline
\end{tabular}
}
\label{Joint annotation PASCAL}
\end{table*}

\begin{table}[t]
  \caption{Our human part definition in PASCAL VOC Persion Part dataset.}
\centering

{
\begin{tabular}{l|l|l|l|l}
\hline
head                                                                                            & \multicolumn{1}{c|}{torso} & arm                                                                  & leg                                                                  & object    \\ \hline
\begin{tabular}[c]{@{}l@{}}hair / head \\ ear / eye \\ eyebrow \\ mouth \\ neck / nose\end{tabular} & torso                      & \begin{tabular}[c]{@{}l@{}}lower arm\\ upper arm \\ hand\end{tabular} & \begin{tabular}[c]{@{}l@{}}lower leg\\ upper leg \\ foot\end{tabular} & all parts \\ \hline
\end{tabular}
}
\label{part dataset}
\end{table}

We generate refined pseudo masks by minimizing the above energy function using graph cut again, and train a new FCN model using the refined pseudo masks.
The whole process is illustrated in Fig. \ref{iter_train}.
In the first iteration, we only use object keypoints to generate pseudo masks and then train the FCN.
From the second iteration, we use object keypoints together with FCN part prediction score map from the last iteration to generate
new pseudo masks. We repeat these steps for a few iterations.
The FCN model in the last iteration is the final model for producing part segmentation prediction.

\subsection{Correlation network for joint prediction}
\label{Correlation network for joint prediction}
There is a strong correlation between part and object segmentation.
It is expected joint part and object segmentation can benefit each other.
In our FCN model, we propose a correlation block to formulate the interaction between parts and objects.
Usually, it is easier to segment out an object correctly than segmenting a part.
Thus, we propose to use object information to guide part prediction.
Fig. \ref{structure loss} shows the framework of the joint inference.

In this module, we generate foreground and background probability for all spatial locations from the object score map.
Then we perform element-wise multiplication between the probability map of objects and the part score map to generate a refined part score map.
In the training step, we add a dense classification loss to this refined part score map for training.
The loss function can be formulated as:

\begin{align}
L_{part} = \sum_{i = 1}^{n} (-log(P(Z^P_i \mid X) \otimes P(Z^O_i \mid X))) ,
\end{align}
where $ Z_i^P $ is the part prediction from the network and $ Z_i^O $ is the object
prediction. The symbol $ \otimes $ indicates element-wise multiplication.
Here $ Z^P_i $ with $K+1$ dimensions contains the output of $ K $ parts and the background category. $ Z_i^O $ contains the foreground probability and the background probability where we repeat the foreground probability $ K $ times to match the dimension of $ Z^P_i $.

\begin{table}[t]
  \caption{Result comparison (IoU scores) between a fully supervised method with pixel-level annotations and our weakly supervised method with object keypoint annotations.}
\centering

\resizebox{1\linewidth}{!}{
\begin{tabular}{|l|llllll|l|}
\hline
\multirow{2}{*}{}                                                        & \multicolumn{6}{c|}{Part}                     & \multirow{2}{*}{Object} \\ \cline{2-7}
                                                                         & head  & torso & arm   & leg   & bg    & mean  &                         \\ \hline
\begin{tabular}[c]{@{}l@{}}Weakly Supervised\\ (ours only Graph Cut)\end{tabular}       &48.82 & 33.41 & 34.11 & 32.21 &83.81  &46.47  & 52.72                   \\ \hline
\begin{tabular}[c]{@{}l@{}}Weakly Supervised\\ (ours VGG)\end{tabular}       & 55.85 & 35.65 & 27.97 & 25.34 & 87.73  & 46.50 & 58.26                   \\ \hline
\begin{tabular}[c]{@{}l@{}}Weakly Supervised\\ (ours ResNet)\end{tabular}       &  55.79 & 40.59 & 32.63 & 37.98 & 87.40 & 50.86  & 59.82                  \\ \hline
\begin{tabular}[c]{@{}l@{}}Fully Supervised\\ (upper bound)\end{tabular} & 66.47 & 47.82 & 39.93 & 34.24 & 92.79 & 56.25 & 70.29                   \\ \hline
\end{tabular}
}
\label{gap_GT}
\end{table}

\begin{table*}[t]
  \caption{Results for using keypoint annotations in the test time.
  Training column indicates the type of keypoint annotations used for training.
  Testing column indicates the type of keypoint annotations used in the test time.
  The segmentation results are improved for incorporating keypoints in the test time using graph cut.
  }
   \centering
  \resizebox{.8\linewidth}{!}{
  \begin{tabular}{|l|l|llllll|l|}
  \hline
  \multirow{2}{*}{}                       &    & \multicolumn{6}{c|}{Part }                  & \multirow{2}{*}{Object } \\ \cline{3-8}
  Training           & Testing   & head  & torso & arm   & leg   & bg    & mean  &                                \\ \hline \hline

Human Pose dataset                           &  Human Pose dataset  & 58.72 & 39.89 & 35.16 & 33.87 & 87.86 & 51.10 & 60.84                          \\ \hline
  Human Pose dataset                        &   Mask RCNN  & 56.42 & 37.87 & 34.21 & 30.47 & 87.80 & 49.36 & 58.90                          \\ \hline
  Human Pose dataset                        & AlphaPose   & 56.41 & 38.88 & 34.03 & 32.24 & 87.63 & 49.84 & 58.66                          \\ \hline \hline

  Mask RCNN      &  Mask RCNN   & 47.79 & 38.09 & 33.96 & 30.24 & 87.23 & 47.46 & 56.78                          \\ \hline \hline

  AlphaPose      &   AlphaPose & 47.25 & 37.94 & 33.87 & 30.37 & 87.12 & 47.31 & 56.09                          \\ \hline
  \end{tabular}
  }

    \label{fourth graph cut}
\end{table*}

\begin{table}[t]
  \caption{Ablation study of our iterative refinement for pseudo mask generation.
  The part segmentation results shown below are generated by the FCN based segmentation network trained on the generated pseudo masks. Results are IoU scores.
}
\centering
\resizebox{1\linewidth}{!}{
\label{iteration joint annotation}
\begin{tabular}{|l|llllll|l|}
\hline
\multirow{2}{*}{Iter.} & \multicolumn{6}{c|}{Part}                           & \multirow{2}{*}{Object} \\ \cline{2-7}
                           & head  & torso & arm   & leg   & bg     & mean       &                         \\ \hline
1                          & 50.95 & 32.71 & 26.33 & 23.53 & 86.44  & 43.99      & 53.00                   \\ \hline
2                          & 54.35 & 35.58 & 27.95 & 25.71 & 87.24  & 46.17      & 56.70                   \\ \hline
3                          & 55.19 & 35.72 & 28.02 & 25.58 & 87.46  & 46.39      & 57.38                   \\ \hline
4                            &55.70  & 35.67 & 27.88 & 25.62 & 87.54  & 46.48      & 57.68                   \\ \hline
5                          & 55.85 & 35.65 & 27.97 & 25.34 & 87.73  & 46.50      & 58.26                   \\ \hline
\end{tabular}
}
\end{table}

\begin{table}[t]
  \caption{Ablation study of our correlation network for joint learning of objects and parts on PASCAL Human Part dataset. ``Part loss only" means the part prediction only uses the part segmentation branch.
  Results are IoU scores.
  }
\centering

\resizebox{1\linewidth}{!}{
\begin{tabular}{|l|l|l|l|l|l|l|l|}
\hline
                             & \multicolumn{1}{c|}{head} & torso  & arm    & leg    & bg     & mean \\ \hline
Part loss only                           & 49.47         & 32.19  & 25.36  & 22.78  & 86.08  & 43.18 \\ \hline
Joint learning               & 50.95                     & 32.71  & 26.33  & 23.53  & 86.44  & 43.99        \\ \hline
\end{tabular}
}
\label{correction loss}
\end{table}

\begin{table}[t]
  \caption{Results of using different types of keypoint annotations for learning.
  Results are IoU scores on PASCAL Human Part dataset.
  }
    \centering
    \resizebox{1\linewidth}{!}{
  \begin{tabular}{|l|llllll|l|}
  \hline
  \multirow{2}{*}{}                           & \multicolumn{6}{c|}{Part }                  & \multirow{2}{*}{Object } \\ \cline{2-7}
  Annotation              & head  & torso & arm   & leg   & bg    & mean  &                                \\ \hline
  Pose dataset                          & 55.19 & 35.72 & 28.02 & 25.58 & 87.46  & 46.39      & 57.38                   \\ \hline
  Mask RCNN                                 & 44.91 & 34.72 & 26.49 & 24.04 & 86.75 & 43.38 & 53.88                          \\ \hline
  AlphaPose                             & 43.87 & 34.30 & 27.20 & 24.83 & 86.54 & 43.35 & 53.51                          \\ \hline

  \end{tabular}
}
    \label{pre-trained keypoint}
\end{table}

\section{Experiments}

\label{4.1}

We use the PASCAL VOC Person Part dataset to evaluate our weakly supervised method.

We merge some fine level parts in Person Part dataset to match with our defined part categories based on the keypoint annotations. We focus on four types of human parts: head, torso, arm, and leg. The detailed merge strategy can be found in Table \ref{part dataset}.

We use object keypoints for learning our weakly supervised method.
Object keypoints can be obtained from PASCAL VOC Human Pose dataset \cite{xia2017joint}, or from pre-trained object keypoints detector such as Mask RCNN \cite{he2017mask}
and AlphaPose \cite{fang2017rmpe,2018arXiv180200977X}.
If not specifically mentioned, we use the keypoint annotations from the PASCAL VOC Human Pose dataset for training.

\subsection{Implementation details}
\label{implement details}

In the pseudo mask generation step, we generate the initial confident foreground and background regions from keypoint annotations for constructing the unary item in the graph cut model. For the initial foreground regions, we employ the labeling strategy in Table \ref{Joint annotation PASCAL} to generate part labels from the keypoint annotations and keypoint connections. For the initial background regions, we set the regions which are at least $50$ pixels away from the nearest keypoint as background regions. The graph-cut optimization step is sensitive to the granularity of the super-pixel. We set the minimum component size as 60.

We trained FCN based segmentation networks using the generated pseudo masks.
Our FCN model is based on the DeepLab method \cite{CY2016Attention}, and we use VGG-16 \cite{DBLP:journals/corr/SimonyanZ14a} as our backbone network.
We set equal weights for the object loss, the part loss and the refined part loss.
We set the batch size to 12 and run 8000 training iterations.
We set the learning rate to 0.001 and reduce the learning rate after every 1000 iteration by a factor of 0.5. The momentum is set to 0.9 and the weigh decay is set to 0.0005.

\subsection{Comparison with fully supervised learning}

\begin{figure*}[t]
\begin{center}

\includegraphics[height=\textheight]{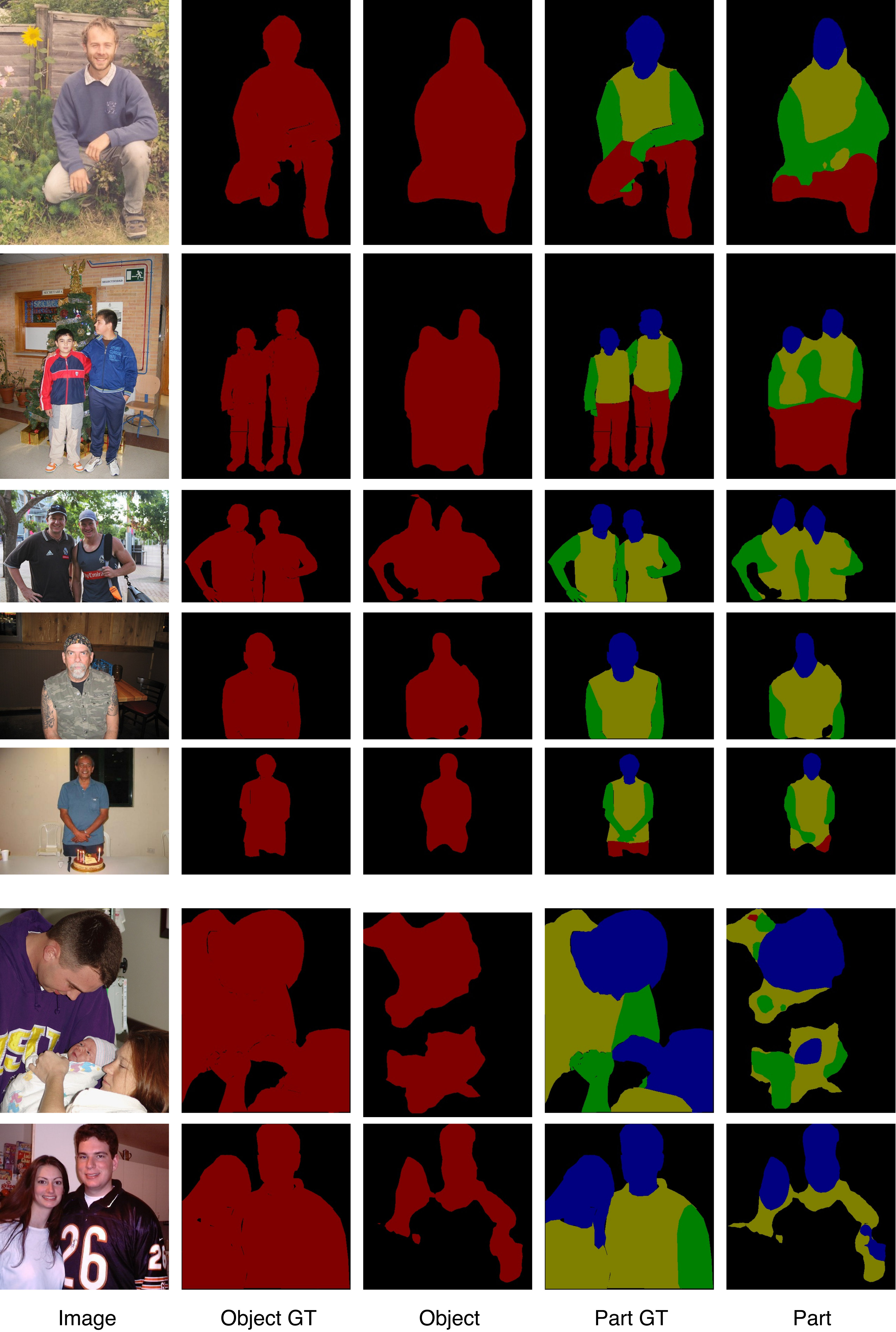}
\caption{Some examples of our weakly supervised human parsing on PASCAL Human Part dataset.
The first five rows show good cases and the last two rows show failure cases.}
\label{pascal_example}
\end{center}
\end{figure*}

We compare the performance of our weakly supervised learning method with the conventional fully supervised method. As shown in Fig. \ref{weakly}, fully supervised part segmentation methods require expensive pixel-level annotations, while our weakly supervised method only uses object keypoint annotations.
The results are shown in Table \ref{gap_GT}. Our weakly supervised method is able to achieve a comparable result with a fully supervised method, with about 10 \% performance drop. Note that the fully supervised method uses the VGG network, same as our VGG based model.
Some prediction examples are shown in Fig. \ref{pascal_example}.

\subsection{Ablation studies}

\textbf{Iterative refinement.}
Table \ref{iteration joint annotation} shows the IoU scores with different numbers of iterations for the pseudo mask refinement. We can see that our iterative refinement approach effectively improves the segmentation. More iterations remarkably improve the final part segmentation performance and it converges after 3 iterations.

\textbf{Object-part joint learning.}
We develop a correlation network for a joint part and object learning.
Table \ref{correction loss} shows the results with and without our correlation block, which demonstrates that our correlation block successfully improves the IoU scores for part segmentation.

\textbf{Using keypoint detectors for learning.}
We evaluate using different types of keypoint detectors to generate object keypoints for our weakly supervised method including the well-known Mask RCNN based keypoint detection method \cite{he2017mask} and AlphaPose method \cite{fang2017rmpe,2018arXiv180200977X}.
We report the part segmentation results in terms of IoU in Table \ref{pre-trained keypoint}.
It shows that our method is able to incorporate less accurate keypoint annotations generated by keypoint detectors for learning, while still able to achieve very competitive part segmentation results.

\textbf{Using keypoints in test time.}
We are able to incorporate the keypoints in the test time to further improve our part segmentation results.
We use Graph Cut optimization \ref{Generate pseudo mask from keypoints} to jointly consider FCN segmentation results and object keypoints to improve the segmentation results, same as that done in training for the pseudo mark refinement. The results are shown in Table \ref{fourth graph cut}.
We choose the well known Mask RCNN \cite{he2017mask} and AlphaPose \cite{fang2017rmpe,2018arXiv180200977X}
as our keypoint detectors.
The results show that using the keypoints in the test time is able to significantly improve the part segmentation performance.

\section{Conclusions}

We have proposed a novel weakly supervised human parsing method which only uses object keypoint annotation for learning. Our method significantly reduces human labeling efforts for pixel-level human parsing tasks.
Particularly, we have developed an iterative learning approach to generate accurate pseudo masks of parts, and we have also developed a correlation network for joint learning of parts and objects, which improves the part segmentation. Our comprehensive ablation study and performance evaluation have justified the effectiveness and usefulness of the proposed method for human parsing.

\clearpage

\bibliography{egbib.bib}

\begin{thebibliography}{}

\bibitem[\protect\citeauthoryear{Bearman \bgroup et al\mbox.\egroup
  }{2016}]{bearman2016s}
Bearman, A.; Russakovsky, O.; Ferrari, V.; and Fei-Fei, L.
\newblock 2016.
\newblock What’s the point: Semantic segmentation with point supervision.
\newblock In {\em European Conference on Computer Vision},  549--565.
\newblock Springer.

\bibitem[\protect\citeauthoryear{Boykov and
  Kolmogorov}{2004}]{boykov2004experimental}
Boykov, Y., and Kolmogorov, V.
\newblock 2004.
\newblock An experimental comparison of min-cut/max-flow algorithms for energy
  minimization in vision.
\newblock {\em IEEE transactions on pattern analysis and machine intelligence}
  26(9):1124--1137.

\bibitem[\protect\citeauthoryear{Cao \bgroup et al\mbox.\egroup
  }{2017}]{cao2017realtime}
Cao, Z.; Simon, T.; Wei, S.-E.; and Sheikh, Y.
\newblock 2017.
\newblock Realtime multi-person 2d pose estimation using part affinity fields.
\newblock In {\em CVPR}.

\bibitem[\protect\citeauthoryear{Chen \bgroup et al\mbox.\egroup
  }{2014}]{chen_cvpr14}
Chen, X.; Mottaghi, R.; Liu, X.; Fidler, S.; Urtasun, R.; and Yuille, A.
\newblock 2014.
\newblock Detect what you can: Detecting and representing objects using
  holistic models and body parts.
\newblock In {\em IEEE Conference on Computer Vision and Pattern Recognition
  (CVPR)}.

\bibitem[\protect\citeauthoryear{Chen \bgroup et al\mbox.\egroup
  }{2016a}]{chen2016deeplab}
Chen, L.-C.; Papandreou, G.; Kokkinos, I.; Murphy, K.; and Yuille, A.~L.
\newblock 2016a.
\newblock Deeplab: Semantic image segmentation with deep convolutional nets,
  atrous convolution, and fully connected crfs.
\newblock {\em arXiv preprint arXiv:1606.00915}.

\bibitem[\protect\citeauthoryear{Chen \bgroup et al\mbox.\egroup
  }{2016b}]{CY2016Attention}
Chen, L.-C.; Yang, Y.; Wang, J.; Xu, W.; and Yuille, A.~L.
\newblock 2016b.
\newblock Attention to scale: Scale-aware semantic image segmentation.
\newblock In {\em CVPR}.

\bibitem[\protect\citeauthoryear{Dai, He, and Sun}{2015}]{dai2015boxsup}
Dai, J.; He, K.; and Sun, J.
\newblock 2015.
\newblock Boxsup: Exploiting bounding boxes to supervise convolutional networks
  for semantic segmentation.
\newblock In {\em Proceedings of the IEEE International Conference on Computer
  Vision},  1635--1643.

\bibitem[\protect\citeauthoryear{Fang \bgroup et al\mbox.\egroup
  }{2017}]{fang2017rmpe}
Fang, H.-S.; Xie, S.; Tai, Y.-W.; and Lu, C.
\newblock 2017.
\newblock {RMPE}: Regional multi-person pose estimation.
\newblock In {\em ICCV}.

\bibitem[\protect\citeauthoryear{He \bgroup et al\mbox.\egroup
  }{2016}]{he2016deep}
He, K.; Zhang, X.; Ren, S.; and Sun, J.
\newblock 2016.
\newblock Deep residual learning for image recognition.
\newblock In {\em Proceedings of the IEEE conference on computer vision and
  pattern recognition},  770--778.

\bibitem[\protect\citeauthoryear{He \bgroup et al\mbox.\egroup
  }{2017}]{he2017mask}
He, K.; Gkioxari, G.; Doll{\'a}r, P.; and Girshick, R.
\newblock 2017.
\newblock Mask r-cnn.
\newblock In {\em Computer Vision (ICCV), 2017 IEEE International Conference
  on},  2980--2988.
\newblock IEEE.

\bibitem[\protect\citeauthoryear{Li \bgroup et al\mbox.\egroup
  }{2017}]{li2017towards}
Li, J.; Zhao, J.; Wei, Y.; Lang, C.; Li, Y.; and Feng, J.
\newblock 2017.
\newblock Towards real world human parsing: Multiple-human parsing in the wild.
\newblock {\em arXiv preprint arXiv:1705.07206}.

\bibitem[\protect\citeauthoryear{Liang \bgroup et al\mbox.\egroup
  }{2015}]{liang2015deep}
Liang, X.; Liu, S.; Shen, X.; Yang, J.; Liu, L.; Dong, J.; Lin, L.; and Yan, S.
\newblock 2015.
\newblock Deep human parsing with active template regression.
\newblock {\em IEEE transactions on pattern analysis and machine intelligence}
  37(12):2402--2414.

\bibitem[\protect\citeauthoryear{Lin \bgroup et al\mbox.\egroup
  }{2014}]{lin2014microsoft}
Lin, T.-Y.; Maire, M.; Belongie, S.; Hays, J.; Perona, P.; Ramanan, D.;
  Doll{\'a}r, P.; and Zitnick, C.~L.
\newblock 2014.
\newblock Microsoft coco: Common objects in context.
\newblock In {\em European conference on computer vision},  740--755.
\newblock Springer.

\bibitem[\protect\citeauthoryear{Lin \bgroup et al\mbox.\egroup
  }{2016}]{lin2016scribblesup}
Lin, D.; Dai, J.; Jia, J.; He, K.; and Sun, J.
\newblock 2016.
\newblock Scribblesup: Scribble-supervised convolutional networks for semantic
  segmentation.
\newblock In {\em Proceedings of the IEEE Conference on Computer Vision and
  Pattern Recognition},  3159--3167.

\bibitem[\protect\citeauthoryear{Lin \bgroup et al\mbox.\egroup
  }{2017}]{Lin:2017:RefineNet}
Lin, G.; Milan, A.; Shen, C.; and Reid, I.
\newblock 2017.
\newblock Refine{N}et: {M}ulti-path refinement networks for high-resolution
  semantic segmentation.
\newblock In {\em CVPR}.

\bibitem[\protect\citeauthoryear{Long, Shelhamer, and
  Darrell}{2015}]{long2015fully}
Long, J.; Shelhamer, E.; and Darrell, T.
\newblock 2015.
\newblock Fully convolutional models for semantic segmentation.
\newblock In {\em CVPR}, volume~3, ~4.

\bibitem[\protect\citeauthoryear{Papandreou \bgroup et al\mbox.\egroup
  }{2015}]{PC2015Weak}
Papandreou, G.; Chen, L.-C.; Murphy, K.; and Yuille, A.~L.
\newblock 2015.
\newblock Weakly- and semi-supervised learning of a dcnn for semantic image
  segmentation.
\newblock In {\em ICCV}.

\bibitem[\protect\citeauthoryear{Pathak \bgroup et al\mbox.\egroup
  }{2014}]{pathak2014fully}
Pathak, D.; Shelhamer, E.; Long, J.; and Darrell, T.
\newblock 2014.
\newblock Fully convolutional multi-class multiple instance learning.
\newblock {\em arXiv preprint arXiv:1412.7144}.

\bibitem[\protect\citeauthoryear{Shen \bgroup et al\mbox.\egroup
  }{2017}]{ShenBMVC17}
Shen, T.; Lin, G.; Liu, L.; Shen, C.; and Reid, I.~D.
\newblock 2017.
\newblock Weakly supervised semantic segmentation based on co-segmentation.
\newblock In {\em BMVC}.

\bibitem[\protect\citeauthoryear{Simonyan and
  Zisserman}{2014a}]{DBLP:journals/corr/SimonyanZ14a}
Simonyan, K., and Zisserman, A.
\newblock 2014a.
\newblock Very deep convolutional networks for large-scale image recognition.
\newblock {\em CoRR} abs/1409.1556.

\bibitem[\protect\citeauthoryear{Simonyan and
  Zisserman}{2014b}]{simonyan2014very}
Simonyan, K., and Zisserman, A.
\newblock 2014b.
\newblock Very deep convolutional networks for large-scale image recognition.
\newblock {\em arXiv preprint arXiv:1409.1556}.

\bibitem[\protect\citeauthoryear{Tsogkas \bgroup et al\mbox.\egroup
  }{2015}]{tsogkas2015deep}
Tsogkas, S.; Kokkinos, I.; Papandreou, G.; and Vedaldi, A.
\newblock 2015.
\newblock Deep learning for semantic part segmentation with high-level
  guidance.
\newblock {\em arXiv preprint arXiv:1505.02438}.

\bibitem[\protect\citeauthoryear{Uijlings \bgroup et al\mbox.\egroup
  }{2013}]{uijlings2013selective}
Uijlings, J.~R.; Van De~Sande, K.~E.; Gevers, T.; and Smeulders, A.~W.
\newblock 2013.
\newblock Selective search for object recognition.
\newblock {\em International journal of computer vision} 104(2):154--171.

\bibitem[\protect\citeauthoryear{Vernaza and
  Chandraker}{2017}]{vernaza2017learning}
Vernaza, P., and Chandraker, M.
\newblock 2017.
\newblock Learning random-walk label propagation for weakly-supervised semantic
  segmentation.
\newblock In {\em The IEEE Conference on Computer Vision and Pattern
  Recognition (CVPR)}, volume~3.

\bibitem[\protect\citeauthoryear{Wang \bgroup et al\mbox.\egroup
  }{2015}]{wang2015joint}
Wang, P.; Shen, X.; Lin, Z.; Cohen, S.; Price, B.; and Yuille, A.
\newblock 2015.
\newblock Joint object and part segmentation using deep learned potentials.
\newblock {\em ICCV}.

\bibitem[\protect\citeauthoryear{Xia \bgroup et al\mbox.\egroup
  }{2016}]{xia2016zoom}
Xia, F.; Wang, P.; Chen, L.-C.; and Yuille, A.~L.
\newblock 2016.
\newblock Zoom better to see clearer: Human and object parsing with
  hierarchical auto-zoom net.
\newblock In {\em European Conference on Computer Vision},  648--663.
\newblock Springer.

\bibitem[\protect\citeauthoryear{Xia \bgroup et al\mbox.\egroup
  }{2017}]{xia2017joint}
Xia, F.; Wang, P.; Chen, X.; and Yuille, A.
\newblock 2017.
\newblock Joint multi-person pose estimation and semantic part segmentation.
\newblock {\em arXiv preprint arXiv:1708.03383}.

\bibitem[\protect\citeauthoryear{Xiu \bgroup et al\mbox.\egroup
  }{2018}]{2018arXiv180200977X}
Xiu, Y.; Li, J.; Wang, H.; Fang, Y.; and Lu, C.
\newblock 2018.
\newblock {Pose Flow}: Efficient online pose tracking.
\newblock {\em ArXiv e-prints}.

\bibitem[\protect\citeauthoryear{{Zhang} \bgroup et al\mbox.\egroup
  }{2018}]{zhangweakly2018}
{Zhang}, T.; {Lin}, G.; {Cai}, J.; {Shen}, T.; {Shen}, C.; and {Kot}, A.~C.
\newblock 2018.
\newblock {Decoupled Spatial Neural Attention for Weakly Supervised Semantic
  Segmentation}.
\newblock {\em ArXiv e-prints}.

\end{thebibliography}
\bibliographystyle{aaai}

\end{document}